\documentclass[10pt,twocolumn,letterpaper]{article}

\usepackage{iccv}
\usepackage{times}
\usepackage{epsfig}
\usepackage{graphicx}
\usepackage{amsmath}
\usepackage{amssymb}
\usepackage{multirow}
\usepackage{comment}
\usepackage{array}
\newcolumntype{P}[1]{>{\centering\arraybackslash}m{#1}}

\usepackage[pagebackref=true,breaklinks=true,letterpaper=true,colorlinks,bookmarks=false]{hyperref}

\iccvfinalcopy 

\def\iccvPaperID{****} 

\ificcvfinal\fi

\begin{document}

\title{Unravelling the Effect of Image Distortions for Biased Prediction of Pre-trained Face Recognition Models}

\author{Puspita Majumdar$^{1, 2}$ \quad Surbhi Mittal$^{2}$ \quad Richa Singh$^{2}$ \quad Mayank Vatsa$^{2}$\\
$^{1}$IIIT-Delhi, India \quad $^{2}$ IIT Jodhpur, India\\
{\tt\small pushpitam@iiitd.ac.in, \{mittal.5, richa, mvatsa\}@iitj.ac.in}
}

\maketitle
\ificcvfinal\thispagestyle{empty}\fi

\begin{abstract}
Identifying and mitigating bias in deep learning algorithms has gained significant popularity in the past few years due to its impact on the society. Researchers argue that models trained on balanced datasets with good representation provide equal and unbiased performance across subgroups. However, \textit{can seemingly unbiased pre-trained model become biased when input data undergoes certain distortions?} For the first time, we attempt to answer this question in the context of face recognition. We provide a systematic analysis to evaluate the performance of four state-of-the-art deep face recognition models in the presence of image distortions across different \textit{gender} and \textit{race} subgroups. We have observed that image distortions have a relationship with the performance gap of the model across different subgroups.
\end{abstract}

\section{Introduction}
Over the past few years, there has been a growing focus on understanding bias in deep learning models. Researchers have attempted to realize the sources of bias and analyze the performance of pre-trained deep models across different demographic subgroups in face analysis problems (e.g., \textit{male} and \textit{female} are subgroups of gender) \cite{celis2019learning,nagpal2019deep}. It has been shown that human bias incorporated during the collection and curation of data \cite{gordon2013reporting}, and imbalance in training data distribution with respect to a particular subgroup \cite{barocas2016big} are some of the potential sources of bias that lead to unfair predictions. A model performing equally well across different subgroups is considered to be an unbiased model \cite{dwork2012fairness}, while they are considered biased when the model favors one subgroup over the other. In this research, we demonstrate that an initially unbiased model may become biased under certain scenarios such as distortions which raises the doubts on models' robustness. 

Several researchers have analyzed the robustness of deep models under image distortions and designed algorithms to enhance robustness \cite{dodge2016understanding,goswami2018unravelling,grm2017strengths,karahan2016image}. However, none of the studies analyze the effect of distortions on the performance of deep models across different demographic subgroups. With recent incidents of biased prediction of deep models towards particular demographic subgroups \cite{buolamwini2018gender,drozdowski2020demographic}, this analysis is crucial to provide insights towards understanding bias in model prediction.

\begin{figure}[]
\centering
\includegraphics[scale = 0.385]{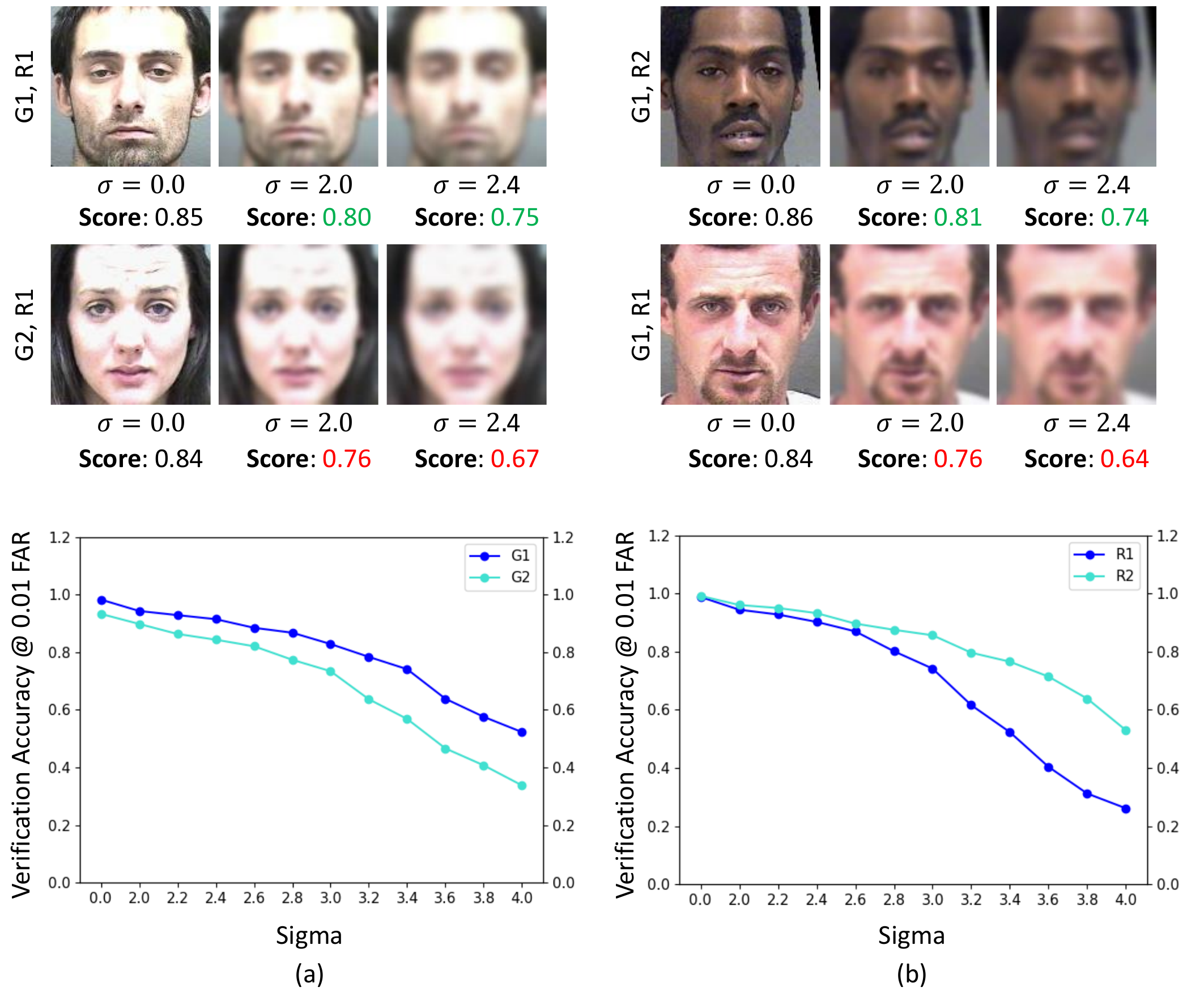}
\caption{Effect of Gaussian blur on the performance of ResNet50 model across different (a) gender and (b) race subgroups. Extracted features of the blurred images are matched with the corresponding clear image using cosine similarity (1.0 is perfect match). Variation in similarity score is shown in the top row. Bottom row shows the verification performance.}
\label{fig:Visual_Abstract}
\end{figure}

As shown in Figure \ref{fig:Visual_Abstract}, experiments conducted to evaluate the performance of face recognition across different subgroups in presence of Gaussian blur show that the confidence of recognizing images is lower for some subgroups compared to others when the same intensity of `blur' is applied. With the objective to unravel this effect with different real world distortions, we investigate two key aspects: \\
\textit{1. Do unbiased model predictions become biased in presence of image distortions? \\
2. How does the performance vary across different demographic subgroups when images undergo distortions?}

We perform a detailed analysis to understand the reason for the difference in the performance of the models across different subgroups in the presence of distortions. Further, we analyze the changes in the regions-of-importance through feature visualization of the salient regions provided by the models. Finally, we discuss some of the possible solutions to overcome the problem of biased prediction of deep models in the presence of image distortions.

\section{Related Work}

This section is segregated into studies related to (i) the effect of image distortions on deep models and (ii) understanding bias.

\subsection{Effect of Image Distortions on Deep Models}
Karahan et al. \cite{karahan2016image} provided a systematic analysis to assess the effect of image quality on face recognition performance using three popular deep models. It is observed that blur, noise, and occlusion significantly deteriorate the performance of deep models. Dodge and Karam \cite{dodge2016understanding} have shown that image distortions result in overall degradation in the classification performance. Grm et al. \cite{grm2017strengths} analyzed the effect of image distortions on face verification performance of four deep models using the LFW dataset \cite{huang2008labeled}. The results indicate that noise, blur, missing pixels, and brightness have a significant effect on the overall model performance. Later, RichardWebster et al. \cite{richardwebster2018visual} studied the behavior of deep models in the presence of image distortions through visual psychophysics and have shown that visual psychophysics makes face recognition more explainable. Two covariate related problems on unconstrained face verification are studied by Lu et al. \cite{lu2019experimental}. First, the effect of covariates is analyzed and then it is utilized to improve verification performance. It is observed that pose variations and occlusions severely affect performance. Also, covariates such as age, gender, and skin tone have shown impacts on performance. Recently, Yang et al. \cite{yang2020advancing} provided a detailed study of object and face detection in poor visibility conditions.

\subsection{Understanding Bias}
In the recent literature, several studies have focused on detection and mitigation of bias present in deep learning models \cite{gong2019debface,majumdar2020subgroup,Wang_2019_ICCV}. However, very few focus on the analysis of this prevalent bias. In \cite{buolamwini2018gender}, the authors unveil the disparity in the performance of commercial-grade gender classification systems based on phenotypic subgroups- lighter-skinned males, darker-skinned males, lighter-skinned females, and darker-skinned females. In \cite{clapes2018apparent}, analysis of real and apparent age differences and its correlation with other attributes is performed. In another work by Nagpal et al. \cite{nagpal2019deep}, the authors attempt to answer how bias is encoded in facial recognition models. They perform analysis based on state-of-the-art deep learning models and suggest how bias encoded by models is comparable to human biases. Wang et al. \cite{wang2019balanced} highlight how balanced datasets are not enough to ensure unbiased performance in deep learning models. They further show how existing biases are amplified by deep learning models in visual recognition tasks. In \cite{celis2019learning}, the authors analyze latent representation of facial images to identify sources of bias. They show that protected attributes like race play a role in biasing latent representations. Recently in \cite{terhorst2020face}, the authors provide an in-depth analysis of the correlation between the quality of face images and biased facial recognition systems. They show that the current definition of face quality transfers bias from face quality assessment systems to facial recognition systems for certain subgroups. In \cite{serna2019algorithmic}, the authors focus on understanding the feature space generated by deep models. Krishnapriya et al. \cite{krishnapriya2020issues} highlight the issues related to face recognition performance with varying skin tone and race. A detailed survey of demographic bias in biometric systems is presented in \cite{drozdowski2020demographic}.

\section{Proposed Evaluation Framework}

In this research, we start with the hypotheses that (i) image distortions affect the performance of a model and cause biased predictions, and (ii) the performance varies differently across different demographic subgroups when images undergo real-world distortions.   
In order to validate and substantiate these hypotheses, the performance of pre-trained deep models is evaluated with and without the presence of distortions across different demographic subgroups. We analyze the regions used by the models for recognition and observe whether the regions of interest remain consistent under the effect of distortions. For this purpose, experiments are performed for face recognition across \textit{gender} and \textit{race} subgroups, using data corresponding to \textit{gender G1 (Male)} and \textit{G2 (Female)}, and \textit{race R1 (light skin color)} and \textit{R2 (dark skin color)}. We have also analyzed the performance of the models across the intersectional subgroups\footnote{The performance of the models across the intersectional subgroups is reported in the supplementary file at: https://github.com/Puspitamajumdariiit/Unravelling-the-Effect-of-Image-Distortions.} of \textit{gender} and \textit{race} - \textit{\{G1,R1\}}, \textit{\{G1,R2\}}, \textit{\{G2,R1\}}, and \textit{\{G2,R2\}}. For the experiments, four pre-trained deep face recognition models: (i) LightCNN-29 \cite{wu2018light} (ii) SENet50 \cite{hu2018squeeze} (iii) ResNet50 \cite{he2016deep}, and (iv) ArcFace \cite{deng2019arcface} are used. Experimental details are shown in Table \ref{Tab: Expt}. The details of the pre-trained models are given in the supplementary file. This section discusses the details of the datasets with the corresponding protocols, image distortions considered in this research, and the evaluation metrics.

\begin{table}[]
\centering
\footnotesize
\caption{Details of the experiments for analyzing the performance of face recognition models across gender and race subgroups under the effect of distortions.}
\label{Tab: Expt}
\renewcommand{\arraystretch}{1.1}
\begin{tabular}{|c|c|l|}
\hline
\textbf{Dataset} & \textbf{Demographic Subgroups}                                                                          & \textbf{Pre-trained Models}                                                   \\ \hline
MORPH            & \begin{tabular}[c]{@{}c@{}}G1, G2, R1, R2, \{G1,R1\},\\ \{G1,R2\}, \{G2,R1\}, \{G2,R2\}\end{tabular} & \begin{tabular}[c]{@{}l@{}}LCNN-29, SENet50,\\ ResNet50, ArcFace\end{tabular} \\ \hline
MUCT             & G1, G2                                                                                                  & \begin{tabular}[c]{@{}l@{}}LCNN-29, SENet50,\\ ResNet50, ArcFace\end{tabular} \\ \hline
\end{tabular}
\end{table}

\subsection{Datasets and Protocols}

\noindent Two publicly available constrained face recognition datasets are used to analyze the effect of image distortions on the performance of pre-trained models across different demographic subgroups. Constrained datasets are considered to solely understand the effect of one type of image distortion on the model performance across different subgroups.

\noindent \textbf{MORPH dataset (Album-2) \cite{rawls2009morph}} contains more than 54K images of 13K subjects. The dataset is pre-labeled with two \textit{gender} subgroups,  \textit{Male (G1)} and  \textit{Female (G2)} and six \textit{race} subgroups,  \textit{White (R1)},  \textit{Black (R2)},  \textit{Hispanic},  \textit{Indian},  \textit{Asian}, and  \textit{Other}. The dataset is imbalanced with respect to different subgroups. Therefore, equal subgroup-wise distribution is ensured during the experiments.

\noindent \textbf{MUCT dataset \cite{milborrow2010muct}} consists of 3,755 images of 131 \textit{male (G1)} and 146 \textit{female (G2)} subjects. For the experiments, equal subgroup-wise distribution is ensured. 

\noindent \textbf{Protocol Details:} For evaluation, face verification is performed and the results are reported at 0.01 False Accept Rate (FAR)\footnote{Important experimental observations are presented in the main paper and the remaining results are summarized in the supplementary file.}. Similar to the LFW dataset \cite{huang2008labeled}, we created 10 disjoint splits of image pairs, each having 300 positive and 300 negative pairs. Here, the positive and negative pairs in each split are created corresponding to each subgroup of a demographic group (e.g. \textit{male} and \textit{female} are the subgroups of the demographic group \textit{gender}). The final evaluation is performed on 12000 pairs with 6000 positive and 6000 negative pairs for each demographic group. Also, for analyzing face recognition performance across \textit{gender} subgroups, race-wise equal distribution is ensured, and vice versa. 



\begin{table}[t]
\centering
\footnotesize
\caption{Verification accuracy and DoB (\%) across different \textit{gender} subgroups under occlusion corresponding to the MORPH dataset. Accuracy of the models degrades significantly on occluding the \textit{eyes}, \textit{nose}, and \textit{mask regions}.}
\label{Tab: Occ_Gender_MORPH}
\renewcommand{\arraystretch}{1.1}
\begin{tabular}{|P{0.5cm}|c|c|c|c|c|}
\hline
                                                                       &     & \textbf{LCNN-29} & \textbf{SENet50} & \textbf{ResNet50} & \textbf{ArcFace} \\ \hline
\multirow{3}{*}{\rotatebox{90}{Eyes}}                                                  & G1  & 98.13            & 70.73            & 52.26             & 96.06            \\ \cline{2-6} 
                                                                       & G2  & 92.00            & 48.43            & 48.10             & 86.10            \\ \cline{2-6} 
                                                                       & DoB & 4.33             & 15.77            & 2.94              & 7.04             \\ \hline
\multirow{3}{*}{\rotatebox{90}{Nose}}                                                  & G1  & 94.23            & 74.56            & 71.11             & 92.40            \\ \cline{2-6} 
                                                                       & G2  & 79.83            & 61.26            & 62.26             & 74.96            \\ \cline{2-6} 
                                                                       & DoB & 10.18            & 9.40             & 6.26              & 12.33            \\ \hline
\multirow{3}{*}{\rotatebox{90}{Mouth}}                                                 & G1  & 99.96            & 96.83            & 97.93             & 99.76            \\ \cline{2-6} 
                                                                       & G2  & 99.36            & 88.50            & 90.50             & 98.10            \\ \cline{2-6} 
                                                                       & DoB & 0.42             & 5.89             & 5.25              & 1.17             \\ \hline
\multirow{3}{*}{\rotatebox{90}{\begin{tabular}[c]{@{}c@{}}Fore-\\ head\end{tabular}}}  & G1  & 99.96            & 94.96            & 95.33             & 99.63            \\ \cline{2-6} 
                                                                       & G2  & 99.53            & 83.30            & 87.70             & 99.06            \\ \cline{2-6} 
                                                                       & DoB & 0.30             & 8.24             & 5.40              & 0.40             \\ \hline
\multirow{3}{*}{\rotatebox{90}{\begin{tabular}[c]{@{}c@{}}Left\\ Cheek\end{tabular}}}  & G1  & 100.00           & 96.30            & 96.90             & 99.80            \\ \cline{2-6} 
                                                                       & G2  & 99.60            & 84.10            & 90.00             & 98.66            \\ \cline{2-6} 
                                                                       & DoB & 0.28             & 8.63             & 4.88              & 0.81             \\ \hline
\multirow{3}{*}{\rotatebox{90}{\begin{tabular}[c]{@{}c@{}}Right\\ Cheek\end{tabular}}} & G1  & 99.96            & 96.67            & 97.36             & 99.86            \\ \cline{2-6} 
                                                                       & G2  & 99.66            & 87.06            & 89.20             & 99.00            \\ \cline{2-6} 
                                                                       & DoB & 0.21             & 6.80             & 5.77              & 0.61             \\ \hline
\multirow{3}{*}{\rotatebox{90}{Mask}}                                                  & G1  & 95.23            & 68.40            & 66.26             & 91.96            \\ \cline{2-6} 
                                                                       & G2  & 82.36            & 54.10            & 37.06             & 69.00            \\ \cline{2-6} 
                                                                       & DoB & 9.10             & 10.11            & 20.65             & 16.24            \\ \hline
\end{tabular}
\end{table}


\subsection{Details of Image Distortions}

\noindent To emulate the real-world scenario of matching unconstrained probe images with constrained gallery images, distortions of different levels/intensities are applied to one of the images in each pair (considering it as probe image) for the verification experiments. The following six image distortions are considered for analysis. Sample images of the MUCT dataset after applying image distortions of different intensities are shown in Figure 1 of the supplementary file.

\noindent \textbf{Occlusion:}
Occluded images are generated by occluding seven facial regions: eyes, nose, mouth, forehead, left cheek, right cheek, and area typically covered by protective face masks. We first detect 68 facial keypoints \cite{rosebrock2017facial} which are then utilized in selecting the region to be occluded.

\noindent \textbf{Gaussian Blur:} 
Images are blurred using Gaussian filters with varying standard deviations $\sigma$. The size of the filter is decided as $2\times \left \lceil{(2\sigma)}\right\rceil+1$. We vary $\sigma$ from 2.0 to 4.0 with a constant step size of 0.2.

\noindent \textbf{Brightness:}
To adjust brightness of images, we apply operations as in \cite{grm2017strengths}. Each image is multiplied by a brightness factor $\beta$ from 1.0 to 3.0 with a constant step size of 0.5 and the values are subsequently clipped to lie between the image pixel intensity range (0,255).

\noindent \textbf{Gaussian Noise:}
To generate images with Gaussian noise, an additive Gaussian noise vector with dimensions equal to the size of the image is used. This vector is generated with values of $\sigma$ varying from 10 to 40, with a step-size of 10.

\noindent \textbf{Salt and Pepper Noise:}
To generate images with salt and pepper noise, an image pixel is set to zero with a probability of $p/2$, or set to 255 with a probability of $p/2$ across all image channels. The value of $p$ is varied from 0.03 to 0.15 with a step size of 0.03.

\noindent \textbf{Resolution:} 
We reduce the resolution of the images using cv2 library \cite{mordvintsev2014opencv} in Python with INTER\_AREA interpolation. The resolutions are varied as $96\times96$, $64\times64$, $48\times48$, $32\times32$, and $28\times28$.


\begin{table}[]
\centering
\footnotesize
\caption{Verification accuracy and DoB (\%) across different \textit{race} subgroups under occlusion corresponding to the MORPH dataset. Occlusion of \textit{nose} region significantly degrades the performance of models for subgroup \textit{R2}.}
\label{Tab: Occ_Race_MORPH}
\renewcommand{\arraystretch}{1.1}
\begin{tabular}{|P{0.5cm}|c|c|c|c|c|}
\hline
                                                                       & \textbf{} & \textbf{LCNN-29} & \textbf{SENet50} & \textbf{ResNet50} & \textbf{ArcFace} \\ \hline
\multirow{3}{*}{\rotatebox{90}{Eyes}}                                                  & R1        & 97.80            & 62.50            & 54.03             & 95.20            \\ \cline{2-6} 
                                                                       & R2        & 93.26            & 62.13            & 51.96             & 90.03            \\ \cline{2-6} 
                                                                       & DoB       & 3.21             & 0.26             & 1.46              & 3.66             \\ \hline
\multirow{3}{*}{\rotatebox{90}{Nose}}                                                  & R1        & 94.80            & 77.16            & 67.66             & 90.46            \\ \cline{2-6} 
                                                                       & R2        & 85.40            & 55.36            & 69.53             & 80.76            \\ \cline{2-6} 
                                                                       & DoB       & 6.65             & 15.41            & 1.32              & 6.86             \\ \hline
\multirow{3}{*}{\rotatebox{90}{Mouth}}                                                 & R1        & 99.96            & 96.23            & 96.96             & 99.66            \\ \cline{2-6} 
                                                                       & R2        & 99.86            & 97.43            & 97.53             & 99.46            \\ \cline{2-6} 
                                                                       & DoB       & 0.07             & 0.85             & 0.40              & 0.14             \\ \hline
\multirow{3}{*}{\rotatebox{90}{\begin{tabular}[c]{@{}c@{}}Fore-\\ head\end{tabular}}}  & R1        & 99.96            & 95.20            & 95.83             & 99.86            \\ \cline{2-6} 
                                                                       & R2        & 99.93            & 95.53            & 95.86             & 99.76            \\ \cline{2-6} 
                                                                       & DoB       & 0.02             & 0.23             & 0.02              & 0.07             \\ \hline
\multirow{3}{*}{\rotatebox{90}{\begin{tabular}[c]{@{}c@{}}Left\\ Cheek\end{tabular}}}  & R1        & 100.00           & 97.00            & 97.33             & 99.83            \\ \cline{2-6} 
                                                                       & R2        & 100.00           & 99.00            & 99.00             & 99.83            \\ \cline{2-6} 
                                                                       & DoB       & 0.00             & 1.41             & 1.18              & 0.00             \\ \hline
\multirow{3}{*}{\rotatebox{90}{\begin{tabular}[c]{@{}c@{}}Right\\ Cheek\end{tabular}}} & R1        & 100.00           & 97.33            & 97.43             & 100.00           \\ \cline{2-6} 
                                                                       & R2        & 99.96            & 98.80            & 98.93             & 99.90            \\ \cline{2-6} 
                                                                       & DoB       & 0.03             & 1.04             & 1.06              & 0.07             \\ \hline
\multirow{3}{*}{\rotatebox{90}{Mask}}                                                  & R1        & 93.23            & 78.33            & 49.53             & 87.76            \\ \cline{2-6} 
                                                                       & R2        & 91.11            & 64.53            & 66.43             & 85.03            \\ \cline{2-6} 
                                                                       & DoB       & 1.50             & 9.76             & 11.95             & 1.93             \\ \hline
\end{tabular}
\end{table}


\subsection{Evaluation Metrics}
To evaluate the effect of distortions on face recognition performance for different subgroups, deep features extracted using pre-trained models are matched using cosine distance. Results are reported in terms of verification accuracy across different subgroups. Further, to measure the bias in model predictions, we use Degree of Bias (DoB) \cite{gong2019debface}, which measures the standard deviation of accuracy (Acc) across different subgroups. It is calculated as:  
\begin{equation}
    DoB = std(Acc_{D_j}) \quad \forall j
\end{equation} 
where, $D_j$ represents a subgroup of a demographic group $D$. High performance gap of the model across different subgroups will result in high $DoB$, indicating higher bias in the model prediction. DoB is commonly used for evaluating bias in face recognition models \cite{gong2019debface, wang2020mitigating}.


\section{Analyzing the Effect of Distortions on Bias in Model Predictions}

In real-world applications of face recognition such as surveillance, an input image undergoes some form of image distortion during acquisition, transmission, and storage. Existing studies have shown that distortions have a significant impact on the performance of deep face recognition models. In this study, we move a step forward and try to find \textit{how pre-trained models perform across different gender and race subgroups under the effect of distortions.} It should be noted that the distortions considered in this research are not added adversarially but occur due to common environmental factors.
\\



\begin{figure}[]
\centering
\includegraphics[scale = 0.40]{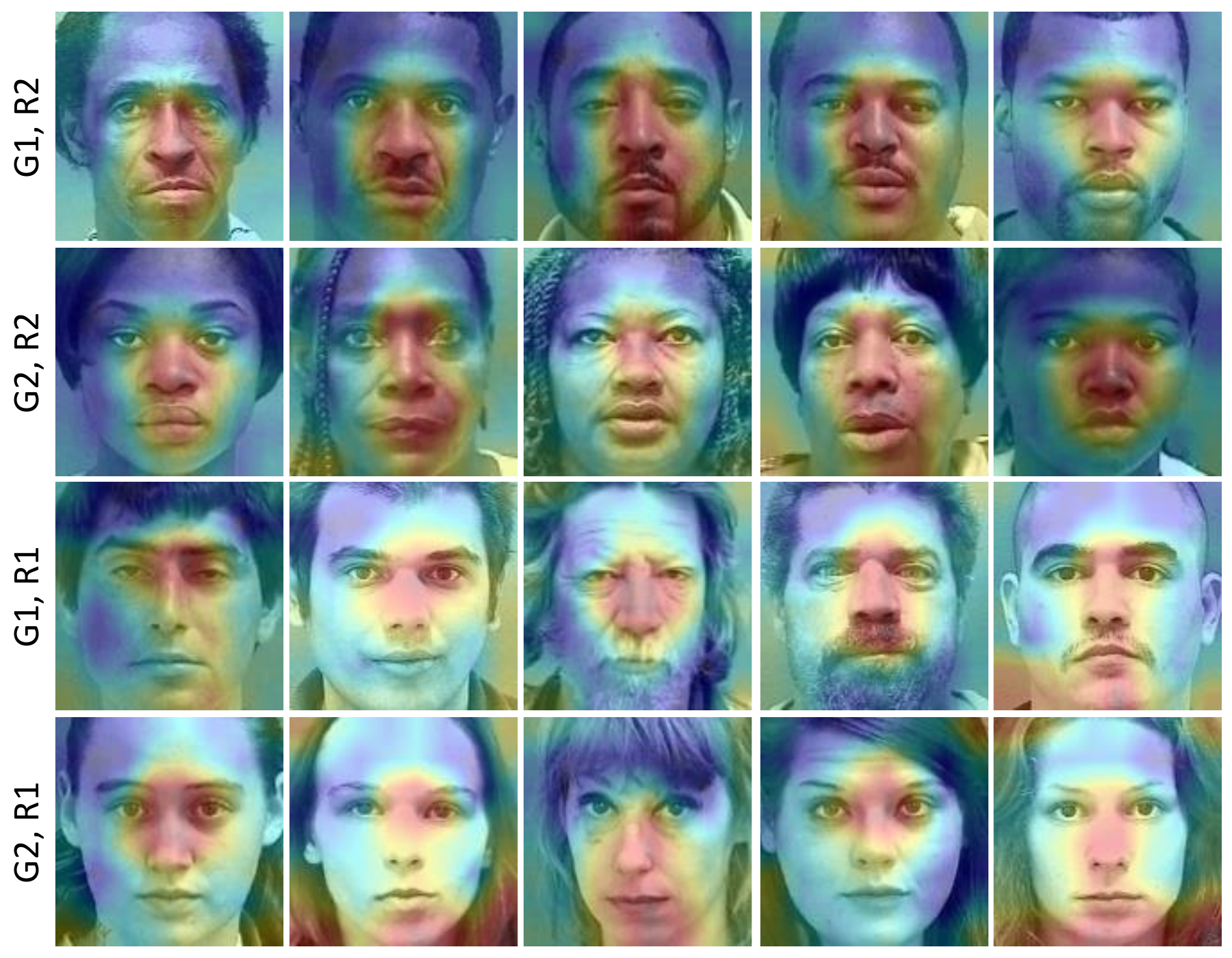}
\caption{Visualization of salient regions of the pre-trained ArcFace model for recognition.}
\label{fig:FeatureMap_Occ}
\end{figure}

\noindent \textbf{Role of facial regions in recognition across subgroups}


We occlude different facial regions to investigate their importance in recognition of a particular subgroup. The verification performance across different \textit{gender} subgroups is shown in Table \ref{Tab: Occ_Gender_MORPH}. The results indicate a significant degradation in performance on occluding the eyes, nose, and facial region covered by a protective face mask. On the other hand, occlusion of mouth, forehead, and cheeks does not have a significant effect. Hence, we can conclude that \textit{eyes, nose, and facial mask regions} are the most discriminative regions for recognition across \textit{gender} subgroups. We also observe that all the models perform poorly for subgroup \textit{G2} resulting in a large performance gap between \textit{G1} and \textit{G2} upon occluding the discriminative regions. 

Similarly, for \textit{race} subgroups (Table \ref{Tab: Occ_Race_MORPH}), eyes, nose, and facial mask regions are found to be the most discriminative regions. It is observed that the difference in performance between \textit{R1} and \textit{R2} is maximum when the nose region is occluded, for most models. The models perform poorly for subgroup \textit{R2} on occluding the nose region. On the other hand, occlusion of other facial regions almost equally affects the performance of the models. \textit{This indicates that nose is the most discriminative region for subgroup \textit{R2}}. To further investigate the underlying reasons for our observation, we have analyzed the regions used by the models for discrimination through feature visualization. The salient regions are obtained by interpolating the final convolution layer filter responses and superimposing on the input image. Figure \ref{fig:FeatureMap_Occ} shows the visualization of salient regions used for feature extraction by the ArcFace model. It is observed that the model focuses predominantly on the eyes and nose regions for feature extraction. For subgroup \textit{R2}, nose is observed to be the most salient region. \\

\begin{table}[]
\centering
\footnotesize
\caption{Verification accuracy and DoB (\%) across different \textit{gender} subgroups with varying intensities of Gaussian blur corresponding to the MORPH dataset. DoB increases with increasing intensities of blur. * represents a relatively high disparity in model performance across different subgroups on undistorted images.}
\label{Tab: GaussianBlur_Iden}
\renewcommand{\arraystretch}{1.1}
\begin{tabular}{|c|c|c|c|c|c|}
\hline
              $\sigma$        &    & \textbf{LCNN-29} & \textbf{SENet50} & \textbf{ResNet50} & \textbf{ArcFace} \\ \hline
\multirow{3}{*}{\rotatebox{90}{0.0}} & G1  & 100.00           & 97.90            & 98.27             & 99.90            \\ \cline{2-6} 
                     & G2  & 99.83            & 91.97            & 93.30             & 99.67            \\ \cline{2-6} 
                     & DoB & 0.12             & 4.19*             & 3.51*              & 0.16             \\ \hline
\multirow{3}{*}{\rotatebox{90}{2.0}} & G1  & 99.83            & 91.23            & 94.30             & 99.70            \\ \cline{2-6} 
                     & G2  & 99.40            & 79.40            & 89.80             & 98.07            \\ \cline{2-6} 
                     & DoB & 0.30             & 8.37             & 3.18              & 1.15             \\ \hline
\multirow{3}{*}{\rotatebox{90}{2.4}} & G1  & 99.73            & 84.53            & 91.47             & 99.30            \\ \cline{2-6} 
                     & G2  & 98.43            & 73.60            & 84.33             & 96.93            \\ \cline{2-6} 
                     & DoB & 0.92             & 7.73             & 5.05              & 1.68             \\ \hline
\multirow{3}{*}{\rotatebox{90}{3.0}} & G1  & 98.83            & 74.00            & 82.83             & 97.07            \\ \cline{2-6} 
                     & G2  & 96.40            & 61.70            & 73.40             & 92.67            \\ \cline{2-6} 
                     & DoB & 1.72             & 8.70             & 6.67              & 3.11             \\ \hline
\multirow{3}{*}{\rotatebox{90}{3.4}} & G1  & 96.87            & 60.83            & 74.10             & 93.70            \\ \cline{2-6} 
                     & G2  & 91.37            & 49.53            & 56.83             & 85.80            \\ \cline{2-6} 
                     & DoB & 3.89             & 7.99             & 12.21             & 5.59             \\ \hline
\multirow{3}{*}{\rotatebox{90}{4.0}} & G1  & 84.70            & 50.57            & 52.27             & 81.07            \\ \cline{2-6} 
                     & G2  & 72.27            & 35.57            & 33.77             & 63.53            \\ \cline{2-6} 
                     & DoB & 8.79             & 10.61            & 13.08             & 12.40            \\ \hline
\end{tabular}
\end{table}

\noindent \textbf{Does model performance degrade equally across subgroups in presence of Gaussian blur?}

Table \ref{Tab: GaussianBlur_Iden}\footnote{We have not reported the results for all $\sigma$ values due to the page limitation. However, a similar trend in the results is observed as shown in the supplementary file.} shows the variation in performance with varying intensities of blur across different \textit{gender} subgroups. It is interesting to observe that an initially unbiased model becomes biased in the presence of blur. The performance gap between \textit{G1} and \textit{G2} increases as we increase the intensity of blur. For instance, the accuracy for \textit{G1} and \textit{G2} is 100\% and 99.83\%, respectively, on original images, corresponding to the LCNN-29 model. However, it reduces to 84.70\% and 72.27\% when degraded with blur with $\sigma=4.0$. As a result, the DoB increases from 0.12\% to 8.79\%, which indicates that bias is introduced in model prediction. For the MUCT dataset, a similar set of observations are drawn regarding the incorporation of bias in model predictions (Table 3 of supplementary file). It is observed that the majority of misclassification occurs in subgroup \textit{G2}. On analyzing the performance across \textit{race} subgroups, we observe that the performance gap increases between \textit{R1} and \textit{R2} with higher performance degradation observed for subgroup \textit{R1} (Table 4 of supplementary file). We have also analyzed the performance across the intersectional subgroups of \textit{gender} and \textit{race} (Table 5 of supplementary file). A huge disparity in model performance across different subgroups is observed.

\begin{table}[]
\centering
\footnotesize
\caption{Verification accuracy and DoB (\%) across different \textit{race} subgroups with varying intensities of brightness corresponding to the MORPH dataset. Accuracy of the models for subgroup \textit{R1} deteriorates significantly in presence of brightness.}
\label{Tab: Brig_Race_MORPH}
\renewcommand{\arraystretch}{1.1}
\begin{tabular}{|c|c|c|c|c|c|}
\hline
$\beta$                    &     & \textbf{LCNN-29} & \textbf{SENet50} & \textbf{ResNet50} & \textbf{ArcFace} \\ \hline
\multirow{3}{*}{\rotatebox{90}{1.0}} & R1  & 100.00           & 98.53            & 98.80             & 99.97            \\ \cline{2-6} 
                     & R2  & 100.00           & 99.27            & 99.17             & 99.93            \\ \cline{2-6} 
                     & DoB & 0.00             & 0.52             & 0.26              & 0.03             \\ \hline
\multirow{3}{*}{\rotatebox{90}{1.5}} & R1  & 99.63            & 70.77            & 28.03             & 91.93            \\ \cline{2-6} 
                     & R2  & 99.83            & 91.93            & 79.30             & 99.40            \\ \cline{2-6} 
                     & DoB & 0.14             & 14.96            & 36.25             & 5.28             \\ \hline
\multirow{3}{*}{\rotatebox{90}{2.0}} & R1  & 87.27            & 26.50            & 2.57              & 43.83            \\ \cline{2-6} 
                     & R2  & 96.37            & 38.23            & 3.80              & 85.50            \\ \cline{2-6} 
                     & DoB & 6.43             & 8.29             & 0.87              & 29.47            \\ \hline
\multirow{3}{*}{\rotatebox{90}{2.5}} & R1  & 52.60            & 11.33            & 1.87              & 16.07            \\ \cline{2-6} 
                     & R2  & 79.40            & 11.70            & 0.93              & 60.07            \\ \cline{2-6} 
                     & DoB & 18.95            & 0.26             & 0.66              & 31.11            \\ \hline
\end{tabular}
\end{table}

\begin{figure}[]
\centering
\includegraphics[scale = 0.33]{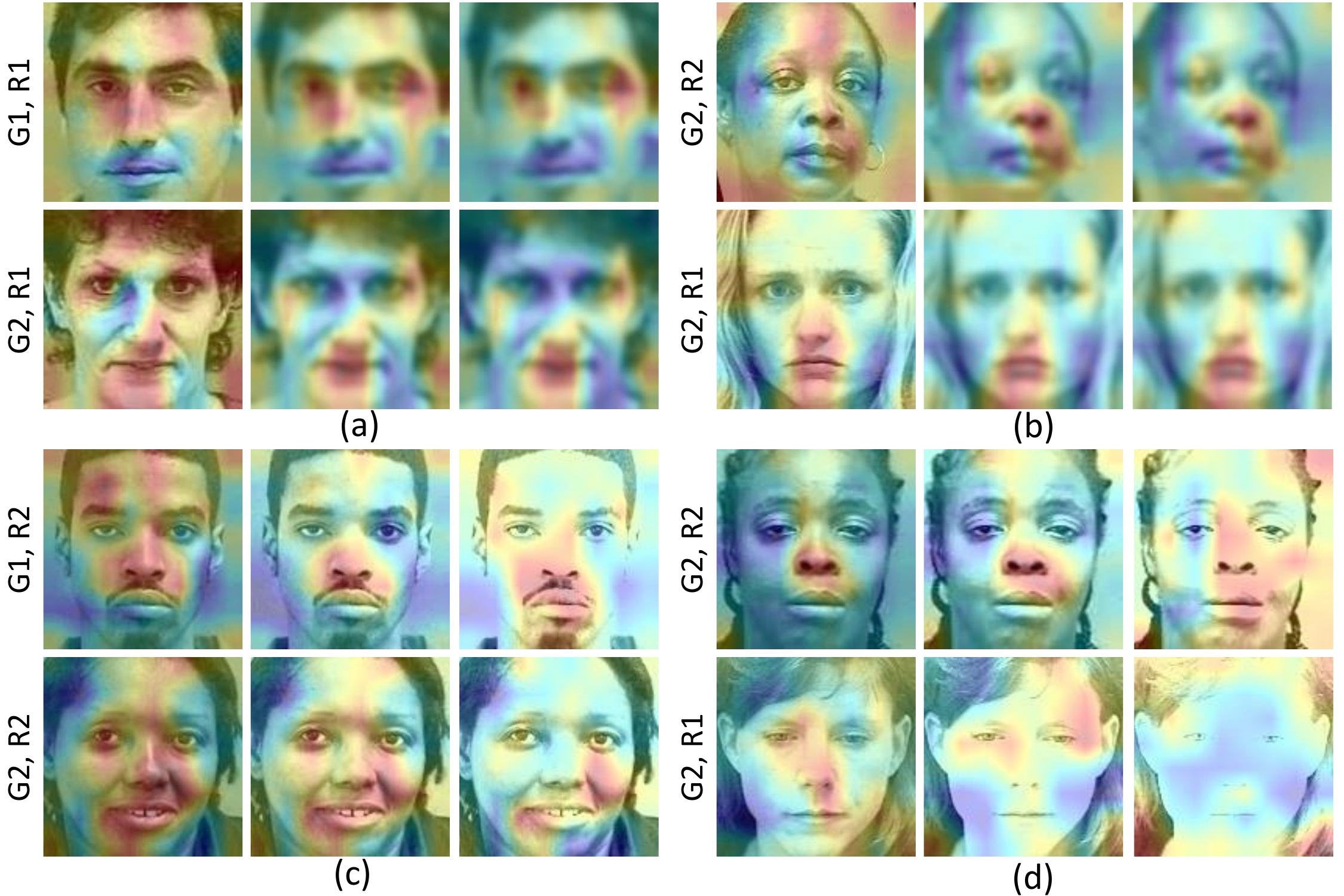}
\caption{Visualizing the salient regions of the pre-trained LCNN-29 model for recognition with varying intensities of (a-b) Gaussian blur and (c-d) brightness. Left block: variation across \textit{gender} subgroups and right block: variation across \textit{race} subgroups. For \textit{gender} subgroups, race is kept constant and vice versa.}
\label{fig:FeatureMap_Blur_Brig}
\end{figure}

To analyze how blur impacts the model's ability to recognize faces across different subgroups, we use feature visualization of salient regions. Figure \ref{fig:FeatureMap_Blur_Brig}(a-b) shows the feature visualization obtained by LCNN-29 on original and blurred images of varying intensities. It is interesting to observe that the regions of interest change when blur is applied. The model shifts the focus from nose to eyes region for subgroup \textit{G1} while it shifts from eyes to mouth region for subgroup \textit{G2}. Among \textit{race} subgroups, the focus shifts from the upper nose region to lower nose region for subgroup \textit{R2}, and to mouth region for subgroup \textit{R1}. As previously seen, mouth region is less discriminative than eyes and nose regions. \textit{This shows that model shifts the regions of interest from higher discriminative regions to lower discriminative regions for subgroups G2 and R1 in presence of blur.} Thus, higher performance degradation is observed for these subgroups.\\



\noindent \textbf{Does performance gap between subgroups increase for brighter images?}

In this experiment, we analyze the effect of \textit{brightness}. The performance is shown across \textit{gender} (Table 6 of supplementary file) and \textit{race} (Table \ref{Tab: Brig_Race_MORPH} of main paper) subgroups of the MORPH dataset. Increasing the intensity of brightness significantly affects the performance of deep models for subgroups \textit{G2} and \textit{R1}. The score distribution corresponding to the SENet50 model shown in Figure 2 of the supplementary file further validates this fact. It is observed that the overlap increases with increasing intensity of brightness. However, subgroups \textit{G1} and \textit{R2} still show more distinct margins compared to \textit{G2} and \textit{R1}, respectively. This indicates that subgroups \textit{G1} and \textit{R2} are still recognizable when exposed to high brightness as compared to \textit{G2} and \textit{R1}, respectively. On analyzing the bias in model prediction, we observe that DoB increases from 4.19\% to 10.25\%, and 0.52\% to 8.29\% for \textit{gender} and \textit{race} subgroups, respectively at $\beta = 2.0$. Similar observations are noted for the MUCT dataset as well (Table 7 of supplementary file).


\begin{table}[]
\centering
\footnotesize
\caption{Verification accuracy and DoB (\%) across different \textit{gender} subgroups with varying intensities of Gaussian noise corresponding to the MORPH dataset. * represents a relatively high disparity in model performance across different subgroups on undistorted images.}
\label{Tab: GNoise_Gender_MORPH}
\renewcommand{\arraystretch}{1.1}
\begin{tabular}{|c|c|c|c|c|c|}
\hline
    $\sigma$                &     & \textbf{LCNN-29} & \textbf{SENet50} & \textbf{ResNet50} & \textbf{ArcFace} \\ \hline
\multirow{3}{*}{\rotatebox{90}{0}}  & G1  & 100.00           & 97.90            & 98.26             & 99.90            \\ \cline{2-6} 
                    & G2  & 99.83            & 91.96            & 93.30             & 99.66            \\ \cline{2-6} 
                    & DoB & 0.12             & 4.19*             & 3.51*              & 0.16             \\ \hline
\multirow{3}{*}{\rotatebox{90}{20}} & G1  & 99.93            & 96.36            & 97.13             & 99.73            \\ \cline{2-6} 
                    & G2  & 99.70            & 90.86            & 89.63             & 98.03            \\ \cline{2-6} 
                    & DoB & 0.16             & 3.89             & 5.30              & 1.20             \\ \hline
\multirow{3}{*}{\rotatebox{90}{30}} & G1  & 99.83            & 94.56            & 93.60             & 99.23            \\ \cline{2-6} 
                    & G2  & 99.36            & 86.46            & 82.80             & 96.66            \\ \cline{2-6} 
                    & DoB & 0.33             & 5.73             & 7.64              & 1.82             \\ \hline
\multirow{3}{*}{\rotatebox{90}{40}} & G1  & 99.63            & 89.93            & 79.40             & 94.10            \\ \cline{2-6} 
                    & G2  & 98.53            & 84.46            & 73.06             & 90.66            \\ \cline{2-6} 
                    & DoB & 0.78             & 3.87             & 4.48              & 2.43             \\ \hline
\end{tabular}
\end{table}


We observe that subgroup \textit{R1} is highly susceptible to the effect of brightness. On increasing the brightness factor, the facial features of subgroup \textit{R1} are heavily affected, which in turn affects the overall performance. The accuracies of all the models for subgroup \textit{R1} drop below 20\% beyond the brightness factor $\beta = 2.5$. We further strengthen the observation using the feature visualization shown in Figure \ref{fig:FeatureMap_Blur_Brig}(d), where we observe that the model is unable to extract features for subgroup \textit{R1} while it focuses on the nose region for subgroup \textit{R2} with increasing intensities of brightness. Similarly, for gender subgroups, the model's focus changes from eyes to nose region for subgroup \textit{G1}, while it shifts from nose to right cheek as shown in Figure \ref{fig:FeatureMap_Blur_Brig}(c). Cheeks are observed to be less discriminative in our occlusion experiment and thus, performance degradation is higher for subgroup \textit{G2}.\\

\noindent \textbf{Do models perform differently across subgroups in presence of noise?}

The effect of \textit{Gaussian Noise} and \textit{Salt and Pepper Noise} are analyzed in the next set of experiments. Table \ref{Tab: GNoise_Gender_MORPH} shows the effect of Gaussian noise on the performance of pre-trained models across different \textit{gender} subgroups. It is observed that the overall performance decreases as the intensity of noise increases, but the performance gap between different subgroups is not significant. For example, the DoB of ResNet50 model increases from 3.51\% to 4.48\% when $\sigma$ is increased to 40. This indicates that model performance is equally affected for \textit{gender} subgroups by Gaussian noise. Similar conclusions are drawn on observing the performance across \textit{race} subgroups (Table 9 of supplementary file). For the MUCT dataset, a similar set of observations are drawn.


\begin{table}[]
\centering
\footnotesize
\caption{Verification accuracy and DoB (\%) across different \textit{gender} subgroups with varying intensities of salt and pepper noise corresponding to the MORPH dataset. The performance of subgroup \textit{G2} gets severely affected with increasing intensities of noise. * represents a relatively high disparity in model performance across different subgroups on undistorted images.}
\label{Tab: SpNoise_Gender_MORPH}
\renewcommand{\arraystretch}{1.1}
\begin{tabular}{|c|c|c|c|c|c|}
\hline
    $p$                  &     & \textbf{LCNN-29} & \textbf{SENet50} & \textbf{ResNet50} & \textbf{ArcFace} \\ \hline
\multirow{3}{*}{\rotatebox{90}{0.00}} & G1  & 100.00           & 97.90            & 98.26             & 99.90            \\ \cline{2-6} 
                      & G2  & 99.83            & 91.96            & 93.30             & 99.66            \\ \cline{2-6} 
                      & DoB & 0.12             & 4.19*             & 3.51*              & 0.16             \\ \hline
\multirow{3}{*}{\rotatebox{90}{0.03}} & G1  & 99.83            & 92.83            & 94.16             & 71.70            \\ \cline{2-6} 
                      & G2  & 98.80            & 82.83            & 83.16             & 55.86            \\ \cline{2-6} 
                      & DoB & 0.73             & 7.07             & 7.78              & 11.20            \\ \hline
\multirow{3}{*}{\rotatebox{90}{0.06}} & G1  & 99.46            & 83.90            & 81.06             & 16.73            \\ \cline{2-6} 
                      & G2  & 96.23            & 72.13            & 63.20             & 10.13            \\ \cline{2-6} 
                      & DoB & 2.28             & 8.32             & 12.63             & 4.67             \\ \hline
\multirow{3}{*}{\rotatebox{90}{0.09}} & G1  & 98.26            & 71.03            & 55.60             & 3.66             \\ \cline{2-6} 
                      & G2  & 94.00            & 58.30            & 35.73             & 3.66             \\ \cline{2-6} 
                      & DoB & 3.01             & 9.00             & 14.05             & 0.00             \\ \hline
\end{tabular}
\end{table}



\begin{figure}[]
\centering
\includegraphics[scale = 0.38]{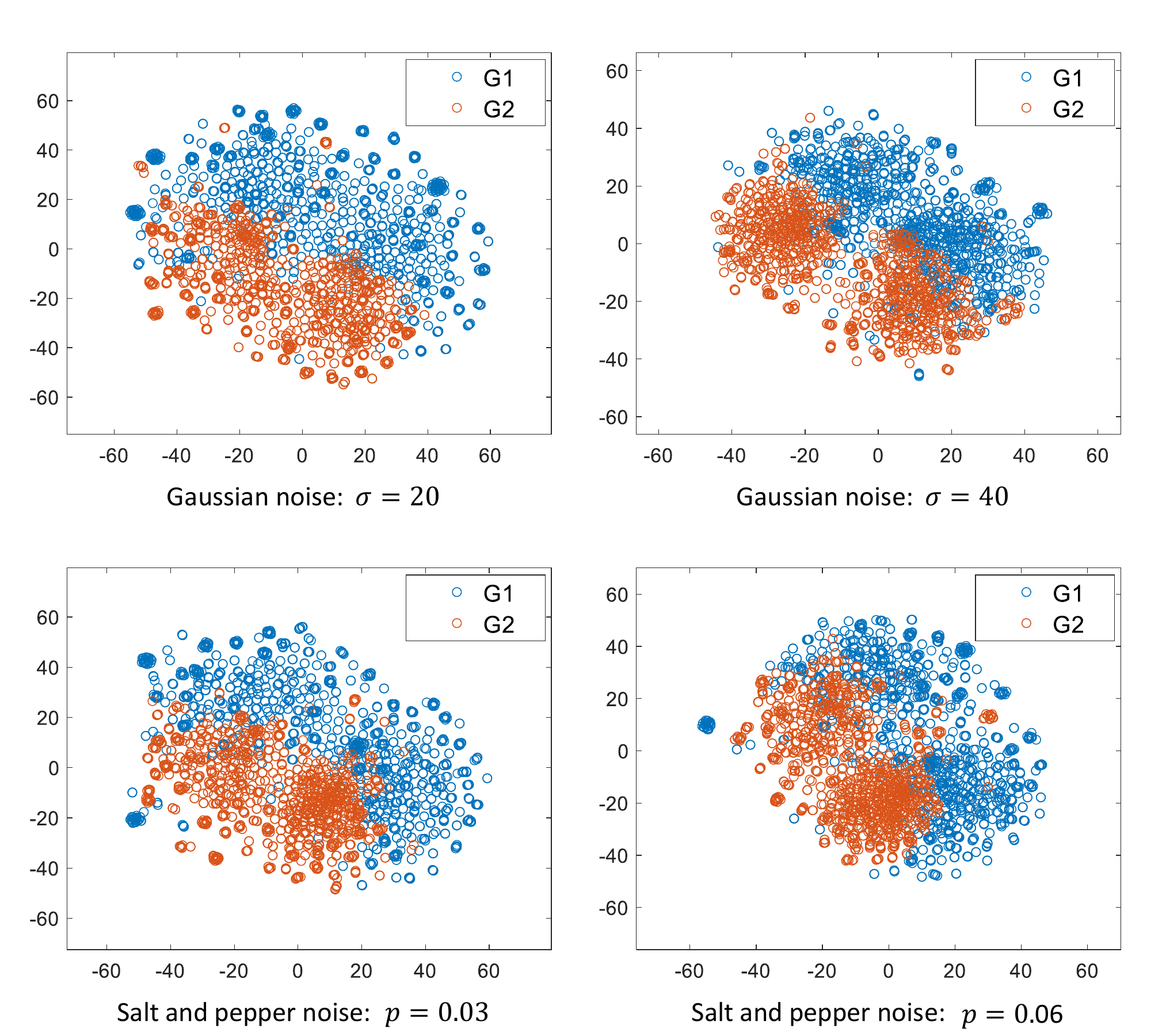}
\caption{t-SNE visualization of ResNet50 features across \textit{gender} subgroups under the effect of Gaussian noise and salt \& pepper noise corresponding to MORPH dataset.}
\label{fig:TSNE_GNoise_SpNoise_Gender}
\end{figure}

The performance of models under the effect of salt and pepper noise across \textit{gender} subgroups is shown in Table \ref{Tab: SpNoise_Gender_MORPH}\footnote{We have reported the values upto $p = 0.09$ due to the page limitation. However, a similar trend in the results is observed for higher values of $p$.}. Here, we observe that unlike the models' performance in presence of Gaussian noise, the performance gap between subgroups increases under the effect of salt and pepper noise. For the ResNet50 model, the DoB increases from 3.51\% to 14.05\%. Similar observations are drawn for \textit{race} subgroups (Table 11 of supplementary file). Earlier studies \cite{karahan2016image,grm2017strengths} have shown that deep models behave similarly for both types of noise. But, in this study, we have observed that salt and pepper noise affect the performance of most of the deep models for subgroups \textit{G2} and \textit{R1} more severely than subgroups \textit{G1} and \textit{R2}, respectively. In order to investigate the difference in behavior of the models for both types of noise across different subgroups, we use the t-SNE visualization for \textit{gender} subgroups, as shown in Figure \ref{fig:TSNE_GNoise_SpNoise_Gender}. For Gaussian noise, it is observed that on increasing the intensity of noise, the overlap in the feature distribution of individual subgroups increases, making the clusters of each subgroup dense. The dense clusters indicate a high misclassification rate and overall performance degradation for each subgroup. On the other hand, with increasing intensities of salt and pepper noise, the cluster of subgroup \textit{G2} becomes denser compared to \textit{G1}. Similar observations are obtained for race subgroups, where the cluster of subgroup \textit{R1} becomes denser compared to \textit{R2}. This indicates high misclassification in subgroups \textit{G2} and \textit{R1}, which in turn affects the performance of these subgroups. 

\begin{figure}[t]
\centering
\includegraphics[scale = 0.33]{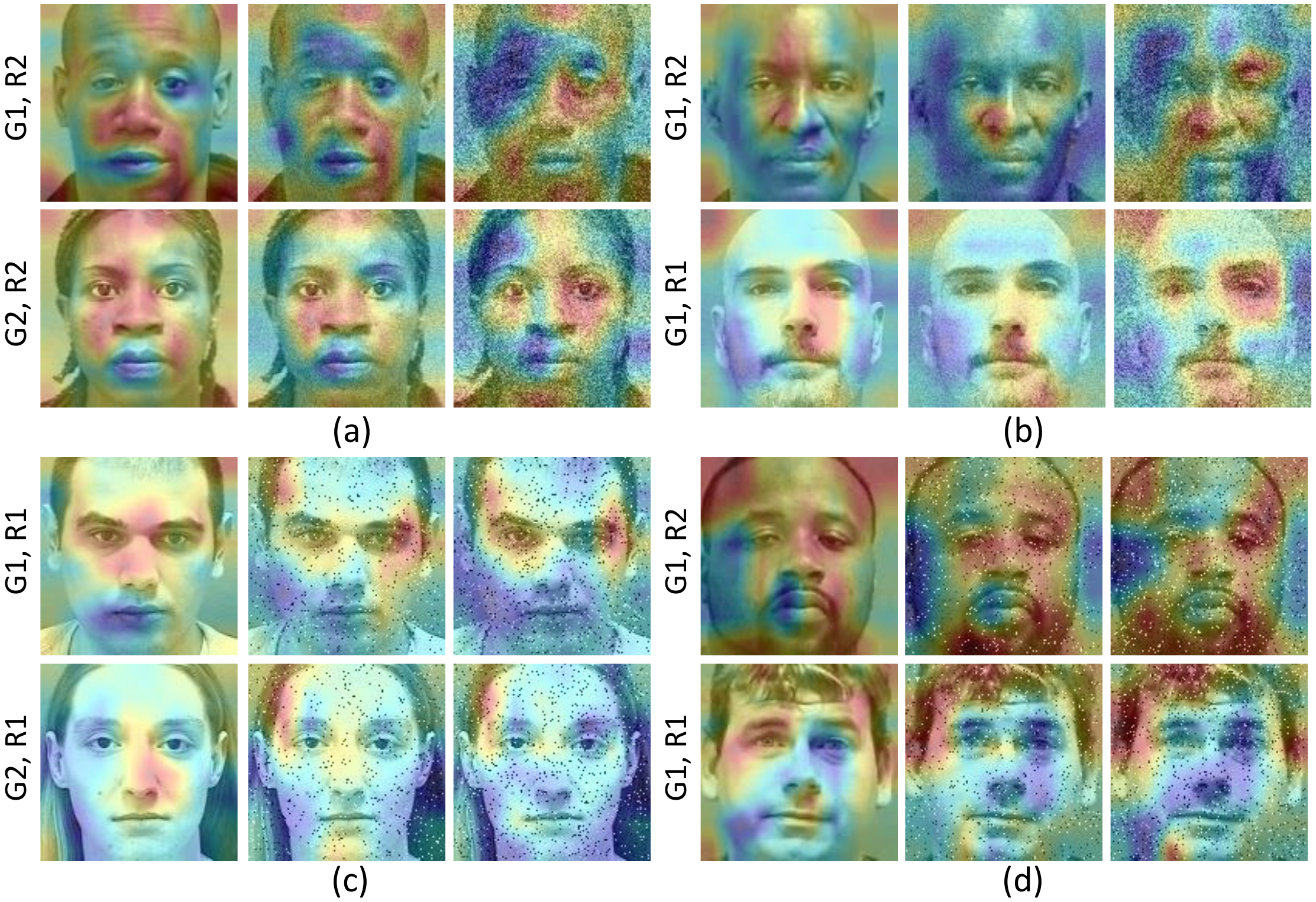}
\caption{Visualization of salient regions used by the pre-trained LCNN-29 model for recognition with varying intensities of (a-b) Gaussian noise and (c-d) salt and pepper noise. Left block - variation across \textit{gender} subgroups and right block - variation across \textit{race} subgroups. For \textit{gender} subgroups, race is kept constant and vice versa.}
\label{fig:FeatureMap_GN_SP}
\end{figure}

We also analyze the salient regions used by the models for recognition under the effect of noise, as shown in Figure \ref{fig:FeatureMap_GN_SP}. It is observed that when Gaussian noise is applied, the region of interest shifts from nose to eyes region. Both these regions are observed to be discriminative in our occlusion experiments, and therefore, minimal performance degradation is observed across different subgroups for the LCNN-29 model. On the other hand, when salt and pepper noise is applied, the model's focus shifts from nose to eyes for subgroup \textit{G1} and to the left side of forehead for subgroup \textit{G2}. Similarly, among \textit{race} subgroups ,the focus changes from eyes to hair for subgroup \textit{R1}, and nose to eyes for subgroup \textit{R2}. As a result, higher performance degradation is observed for \textit{G2} and \textit{R1} in the presence of salt and pepper noise.\\


\begin{table}[t]
\centering
\footnotesize
\caption{Verification accuracy and DoB (\%) across different \textit{gender} subgroups with varying resolution corresponding to the MORPH dataset. A significant degradation in performance is observed at low resolution.}
\label{Tab: Res_Gender_MORPH}
\renewcommand{\arraystretch}{1.1}
\begin{tabular}{|c|c|c|c|c|c|}
\hline
                       &     & \textbf{LCNN-29} & \textbf{SENet50} & \textbf{ResNet50} & \textbf{ArcFace} \\ \hline
\multirow{3}{*}{\rotatebox{90}{$96 \times 96$}} & G1  & 99.97            & 97.57            & 98.33             & 99.90            \\ \cline{2-6} 
                       & G2  & 99.87            & 91.30            & 93.57             & 99.07            \\ \cline{2-6} 
                       & DoB & 0.07             & 4.43             & 3.37              & 0.59             \\ \hline
\multirow{3}{*}{\rotatebox{90}{$64 \times 64$}} & G1  & 99.97            & 96.83            & 97.60             & 99.87            \\ \cline{2-6} 
                       & G2  & 99.80            & 91.50            & 92.53             & 98.70            \\ \cline{2-6} 
                       & DoB & 0.12             & 3.77             & 3.59              & 0.83             \\ \hline
\multirow{3}{*}{\rotatebox{90}{$48 \times 48$}} & G1  & 99.93            & 94.80            & 94.60             & 99.37            \\ \cline{2-6} 
                       & G2  & 99.67            & 87.00            & 90.10             & 95.80            \\ \cline{2-6} 
                       & DoB & 0.18             & 5.52             & 3.18              & 2.52             \\ \hline
\multirow{3}{*}{\rotatebox{90}{$32 \times 32$}} & G1  & 99.70            & 82.87            & 81.40             & 69.43            \\ \cline{2-6} 
                       & G2  & 98.63            & 65.80            & 67.93             & 62.00            \\ \cline{2-6} 
                       & DoB & 0.76             & 12.07            & 9.52              & 5.25             \\ \hline
\multirow{3}{*}{\rotatebox{90}{$28 \times 28$}} & G1  & 98.93            & 65.13            & 63.83             & 9.63             \\ \cline{2-6} 
                       & G2  & 95.17            & 44.50            & 45.30             & 13.87            \\ \cline{2-6} 
                       & DoB & 2.66             & 14.59            & 13.10             & 3.00             \\ \hline
\end{tabular}
\end{table}


\noindent \textbf{Does model performance differ across subgroups with varying image resolution?}

Table \ref{Tab: Res_Gender_MORPH} shows the effect of varying the resolution of the images on the performance of deep models across different \textit{gender} subgroups. A sharp drop in accuracy is observed beyond $48\times48$ resolution for most models. It is also observed that a significant amount of bias is incorporated in predictions of SENet50 and ResNet50 models. As the resolution of the images is reduced to $28\times28$, the DoB reaches upto 14.59\% and 13.10\%, respectively. Similarly, for \textit{race} subgroups, the DoB increases at lower resolutions for SENet50 and ResNet50 models (Table 13 of supplementary file). A huge performance gap of these models is also observed across the intersectional subgroups (Table 14 of supplementary file). On the other hand, a lesser amount of bias is introduced in the LCNN-29 and ArcFace models.

\begin{figure}[]
\centering
\includegraphics[scale = 0.44]{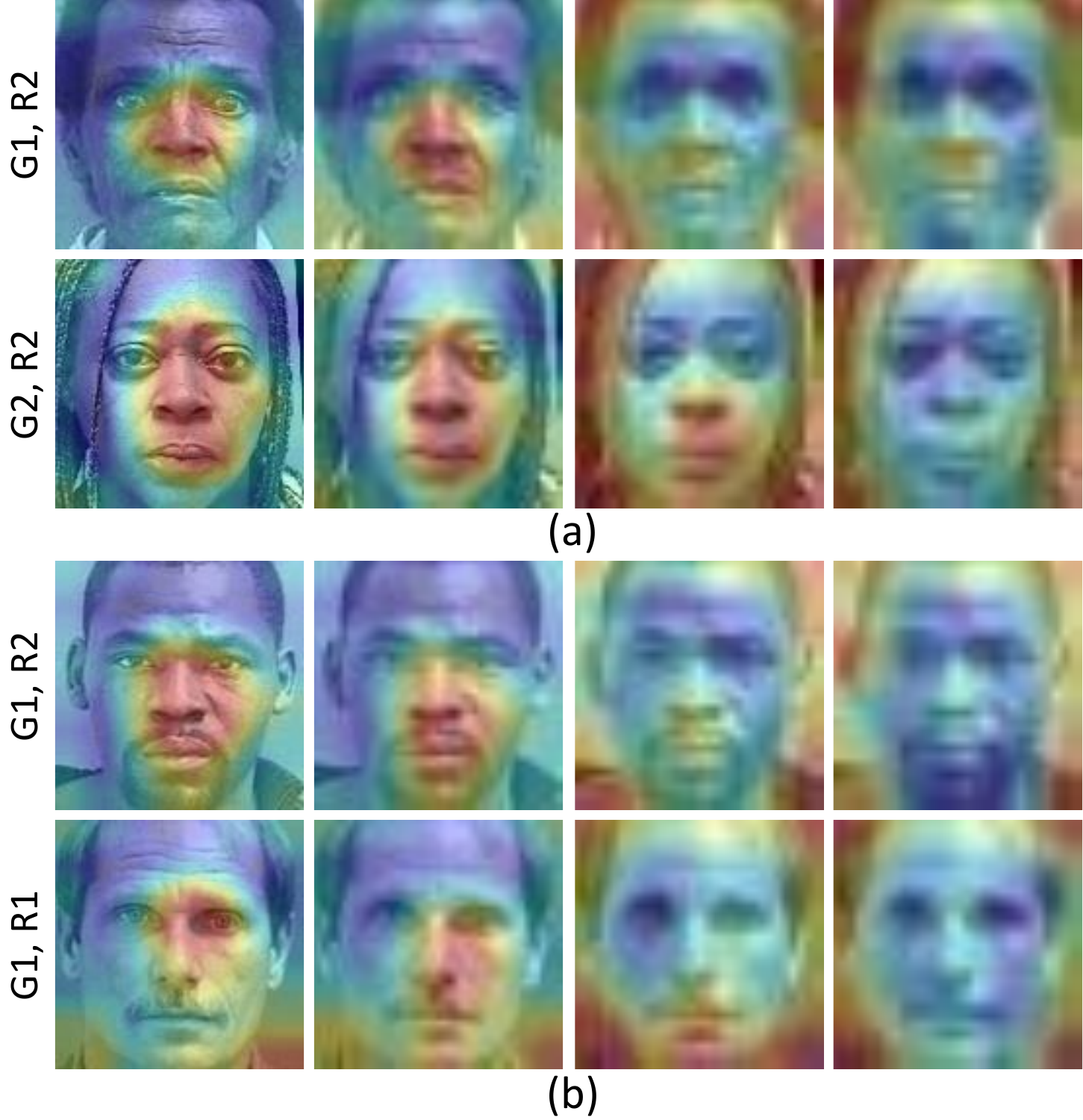}
\caption{Visualization of salient regions used by the pre-trained ArcFace model for recognition with varying resolution across (a) gender and (b) race subgroups. For \textit{gender} subgroups, race is kept constant and vice versa.}
\label{fig:FeatureMap_Res_ArcFace}
\end{figure}

From Table \ref{Tab: Res_Gender_MORPH}, it is observed that ArcFace significantly degrades the performance at low resolution. On varying the resolution of the images from $48\times48$ to $28\times28$, the accuracy of ArcFace drops by 89.74\% and 81.93\% for subgroup \textit{G1} and \textit{G2}, respectively. For the \textit{race} subgroups, it results in 87.73\% and 86.70\% degradation for \textit{R1} and \textit{R2}, respectively. This shows the vulnerability of ArcFace for low resolution image recognition. Figure \ref{fig:FeatureMap_Res_ArcFace} shows the shift in the region of interest under the effect of resolution. It is observed that at low resolution the model is not able to focus on the facial region. Instead, the focus shifts to non-facial regions such as hair, which in turn leads to poor performance. 

\section{Discussion on Unbiased Model Predictions in the Presence of Image Distortions}

The detailed experimental evaluation performed in this research highlights the idea of gender and racial bias as a consequence of degraded image quality. The results also provide important insights and we believe that the observations drawn from the experimental evaluation can open new research threads. To facilitate research in this direction, we discuss some of the possible solutions in the following subsections to ensure reliable and unbiased model predictions in the presence of real-world image distortions.  


\subsection{Image Quality Enhancement}

Enhancing the quality of the images before providing to the face recognition models may reduce the disparity in model predictions. Generative approaches \cite{gu2020image, pan2020exploiting} and denoising techniques \cite{anwar2020identity} can be used to enhance image quality. We assert that matching high-quality images will decrease the performance gap of the model across different subgroups and enhance the overall model performance.

\subsection{Improving Generalizability of Deep Models}

Training deep models for generalized solutions is an important approach for bias mitigation \cite{bahng2020learning, pezeshki2020gradient}. In this context, we believe that training deep face recognition models to extract discriminative features from different facial regions instead of focusing on specific facial regions (e.g., eyes, nose, mouth) for recognition can reduce the bias in model prediction. In other words, different facial regions must be equally discriminative for the models during recognition. In our occlusion experiments, we have observed that the nose is the most discriminative region for recognizing subgroup \textit{R2}. The disparity in the discriminative regions used by models for recognition across different subgroups should be reduced for bias mitigation. 

\subsection{Utilizing Image Quality during Recognition}

The quality of the images should be considered during recognition. In the past, various quality assessment metrics \cite{mittal2012no} and methods \cite{fang2020perceptual, ying2020patches} have been proposed to determine the image quality. Recently, the NTIRE challenge organized in CVPR 2021 focused on perceptual image quality assessment \cite{gu2021ntire}. In the literature, researchers have shown the improvement in recognition performance by utilizing the quality score with model predictions \cite{yu2018face, zhuang2019recognition}. It is our assertion that fusing the quality score with the model prediction will impact the confidence of the model prediction, which may reduce the disparity of the model for recognition across different subgroups.

\section{Conclusion}

This paper analyzes the interplay and effect of bias and real-world image distortions on the performance of face recognition algorithms. The paper contributes in understanding how seemingly unbiased models produce biased predictions in the presence of real-world image distortions. We observe that \textit{eyes}, \textit{nose}, and \textit{mask} are the most discriminative regions for recognition across race and gender subgroups. However, in the presence of distortions, the \textit{regions of interest} used by the models shift towards less discriminative regions, thus resulting in unequal performance degradation. For instance, we observe that the models are biased against gender subgroup \textit{G2} and race subgroup \textit{R1}. Moreover, different models introduce different amounts of bias in the predictions, and they largely favor (or disfavor) the same demographic subgroups. We assert that these understandings are important in building deep learning models that are unbiased under different scenarios.

\section*{Acknowledgements}
P. Majumdar is partly supported by DST Inspire Ph.D. Fellowship. S. Mittal is partially supported by UGC-Net JRF Fellowship. M. Vatsa is partially supported through Swarnajayanti Fellowship. This research is also partially supported by Facebook Ethics in AI award.

{\small
\bibliographystyle{ieee_fullname}
\bibliography{egbib}
}

\end{document}